\documentclass[journal]{IEEEtran}
\usepackage{hyperref}

\usepackage{graphicx}
\ifCLASSINFOpdf
\else
\fi
\usepackage{amssymb}
\usepackage{amsmath}
\usepackage{amsfonts}
\usepackage{amssymb}
\usepackage{booktabs}
\usepackage{multirow}
\usepackage{array}
\usepackage{siunitx}
\usepackage{capt-of}
\usepackage{adjustbox}
\usepackage{array}
\usepackage[numbers]{natbib} 
\usepackage{enumerate}
\usepackage{caption}
\begin{document}
%
\title{The Effectiveness of a Simplified Model Structure\\ for Crowd Counting}
%
%
%

\author{Lei~Chen\IEEEauthorrefmark{2}, 
        Xinghang~Gao\IEEEauthorrefmark{2},
        Fei~Chao,~\IEEEmembership{Member,~IEEE,}
        Xiang~Chang,~\IEEEmembership{Member,~IEEE,}
        Chih-Min Lin,~\IEEEmembership{Fellow,~IEEE,}
        Xingen~Gao\IEEEauthorrefmark{1}, 
        Shaopeng Lin,
        Hongyi Zhang,
        Juqiang Lin
\thanks{This work was supported by Natural Science Foundation of Fujian Province of China (No. 2023J011456 and No.2023J05084).}
\thanks{Lei~Chen, Xinghang~Gao, Xingen~Gao, Shaopeng Lin, Hongyi Zhang, and Juqiang Lin are with the school of Opto-Electronic and Communication Engineering, Xiamen University of Technology, Xiamen, China.}
\thanks{Fei~Chao is with the department of Artificial Intelligence, School of Informatics, Xiamen University, Xiamen, China.}
\thanks{Xiang~Chang is with the department of Computer Science, Institute of Mathematics, Physics and Computer Science, Aberystwyth University, Aberystwyth, UK.}
\thanks{Chih-Min~Lin is with the Department of Electrical Engineering, Yuan Ze University, Taoyuan City 320.}
\thanks{\IEEEauthorrefmark{1} Corresponding author. E-mail: gaoxingen@xmut.edu.cn}
\thanks{\IEEEauthorrefmark{2} These authors contributed equally to the work.}
}

\maketitle

\begin{abstract}
In the field of crowd counting research, many recent deep learning based methods have demonstrated robust capabilities for accurately estimating crowd sizes. However, the enhancement in their performance often arises from an increase in the complexity of the model structure. 
This paper discusses how to construct high-performance crowd counting models using only simple structures. We proposes the Fuss-Free Network (FFNet) that is characterized by its simple and efficieny structure, consisting of only a backbone network and a multi-scale feature fusion structure.  
The multi-scale feature fusion structure is a simple structure consisting of three branches, each only equipped with a focus transition module, and combines the features from these branches through the concatenation operation. 
Our proposed crowd counting model is trained and evaluated on four widely used public datasets, and it achieves accuracy that is comparable to that of existing complex models. 
Furthermore, we conduct a comprehensive evaluation by replacing the existing backbones of various models such as FFNet and CCTrans with different networks, including MobileNet-v3, ConvNeXt-Tiny, and Swin-Transformer-Small. The experimental results further indicate that excellent crowd counting performance can be achieved with the simplied structure proposed by us.
\end{abstract}

\begin{IEEEkeywords}
crowd counting, simplified structure, Fuss-Free Network, multi-scale feature fusion, focus transition module.
\end{IEEEkeywords}

%
\IEEEpeerreviewmaketitle

\section{Introduction}
%
%
%
%
\IEEEPARstart{C}{rowd} counting, which has been receiving growing attention in the field of computer vision~\citep{yan2019perspective}, has the potential to monitor and analyze crowd density in real-time, thereby helping to prevent safety incidents resulting from overcrowding. Moreover, it can provide essential data support for urban management, facilitating the optimization of resource allocation in crowded areas. Additionally, during the pandemic, it has the ability to monitor and regulate crowd movements, contribute to the development of public health strategies, and mitigate the transmission of the virus.

\begin{figure*}[htbp]
    \centering
    \makebox[\textwidth][c]{\includegraphics[width=0.85\textwidth]{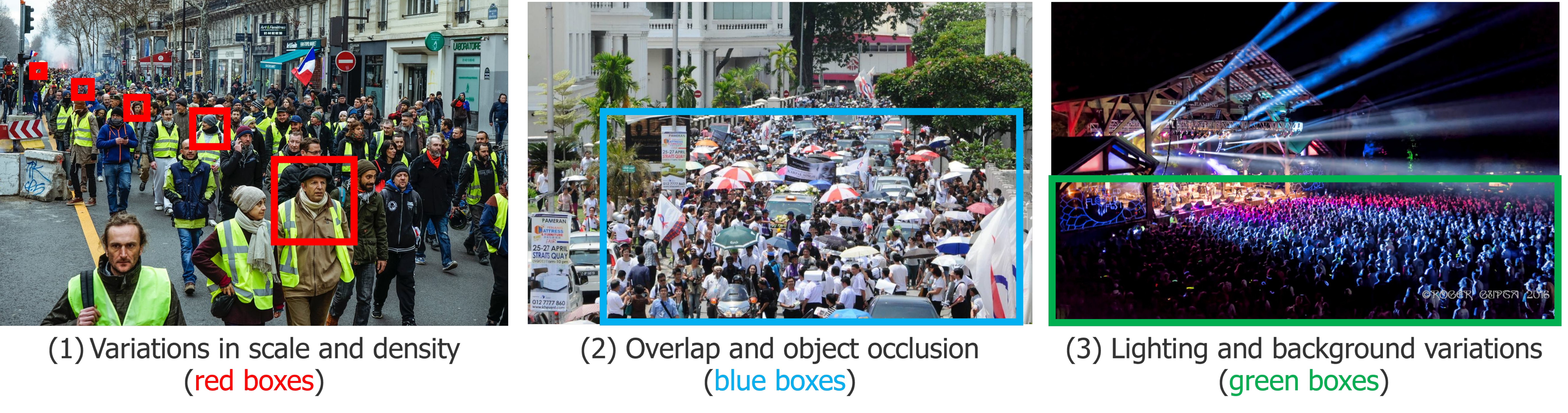}}  
    \caption{Representative images for crowd counting. These images illustrates the main challenges of crowd counting tasks, which include object occlusion and overlap in high density crowds, variations in scale and density within images, as well as the influence of lighting and background discrepancies, among other.}
    \label{fig:crowd_pic}
\end{figure*}

Crowd counting is a highly challenging computer vision task. Figure~\ref{fig:crowd_pic} shows three representative images for crowd counting, illustrating the following challenges in crowd counting tasks: 
(1) Variations in scale and density within an image: In general, crowds that are closer to the camera tend to have larger scales and lower densities, while those further away exhibit the opposite characteristics. 
(2) Overlap occlusion and object in high density crowds: High density crowd counting tasks frequently involve analyzing crowd images where a significant number of individuals are present, and situations such as object occlusion and overlap between individuals are commonly encountered. 
(3) Lighting and background variations: The characteristics of individual features can vary due to differences in lighting conditions, which can include the creation of shadows. Moreover, the complexity and variability of the background environment can introduce interference factors, such as object occlusion. 

To overcome these challenges, mainstream methods of crowd counting rely on the powerful performance of deep learning models. These models are designed and trained to generate crowd density images at the pixel level. Subsequently, an accurate estimation of the total number of individuals is achieved by using a pixel-level crowd density image. 
In order to tackle the issue of scale and density variations in images, many approaches have adopted a multi-branch structure. 
Among them, some methods employ dilated convolutions with different dilation rates in different branchs, enabling the model to have receptive fields of different scales. 
Furthermore, to tackle challenges such as object occlusion, overlap in high density crowds, and lighting and background variations, most methods focus on enhancing the feature extraction capability of the model, primarily by selecting or designing powerful network architectures.
Optimizing the loss function yields significant effectiveness as well. For instance, to enhance model supervision, the loss function can be adjusted to ensure that the model establishes a more precise correspondence between the predicted density and the labeled location.


In this paper, we focus on the impact of structural complexity on modifying models, discussing whether it is possible to find structurally simple models that achieve less performance decrease compared to complex models after modification.
We strive to comprehensively consider the simplicity of the structure and the effectiveness of the model. Therefore, we opt for the ConvNeXT-Tiny architecture as our backbone, renowned for its simplicity and efficient design in convolutional neural networks. 
The multi-scale feature fusion structure we designed is very simple and allows us to effectively utilize information from different scales by fusing features from three different stages within the backbone. To simplify the structure, we only use concatenate operation on the features from three different branches. Moreover, we enhance the model's effectiveness by including a carefully designed focus transition module before the fusion operation. 
In summary, this paper's contributions can be outlined in three aspects:
\begin{enumerate}[(1)]
\item We present a crowd counting model with a simple structure that consists of only a backbone and a multi-scale feature fusion structure. 
Unlike other complex structure models, the fact that modifying our model, such as replacing different backbones, can achieve minimal performance loss demonstrates the effectiveness of this simple structure. 
Additionally, it achieves state-of-the-art (SOTA) results even with low parameters and computational requirements.

\item We investigated three multi-scale feature fusion methods, including concatenate feature fusion, addition feature fusion, and stepwise addition feature fusion. Experimental results demonstrate that concatenate feature fusion has advantages in crowd-counting tasks. It efficiently extracts crowd features and effectively couples them across different scales, enhancing feature representation capability.

\item To effectively extract key information and reduce dimensionality, we have developed a focus transition module that integrates dynamic convolution. It can dynamically adjust multi-dimensional weights based on feature attributes and optimize weights in channel and spatial dimensions.
\end{enumerate}

Early research employed human detection methods~\citep{wu2005detection}, where researchers used detectors to identify humans in an image and then counted each detected person. The drawback of this method is that occlusions between individuals in crowded or  crowded environments greatly affect the performance of detectors, resulting in a decrease in counting accuracy.
Direct regression methods~\citep{chen2012feature} directly estimate the number of people. However, this approach often lacks detailed information about individual distribution and positioning, and it is sensitive to model assumptions, which may impact accuracy in complex or highly variable density scenarios. 
Nowadays, density map regression methods~\citep{zhang2016single} have become popular, as they integrate density map estimates from images to determine the crowd count. These methods rely heavily on strong representation capabilities of networks, with most of them utilizing convolutional neural networks (CNNs). Additionally, some methods also leverage transformers. To enhance performance, researchers have made advancements in technology and architecture. In this section, we review the relevant research on crowd counting, focusing on three aspects: model design, loss function design, and other aspects of the design.

\subsection{Model Design for Crowd Counting}
\label{sec:Model Design for Crowd Counting}

Most of the research focuses on the following three aspects: utilizing powerful backbones, designing multi-scale feature fusion strategies, and integrating attention mechanisms. 

Choosing the right backbone is the primary task for building a crowd counting model. Given the remarkable success of transformer networks in the field of natural language processing (NLP), researchers have also begun to explore their application in crowd counting as a core feature extraction module. Sun et al.~\citep{sun2021boosting} utilize transformer models for feature extraction and information exchange by overlapping and concatenating image blocks and adding contextual tokens. Qian et al.~\citep{qian2022segmentation} introduce a ``U-shaped" multi-scale transformer network for crowd counting and introduce a new loss function to supervise regression, significantly improving counting accuracy.

Multi-scale feature fusion methods attempt to address the challenge of large variations in crowd scale or size diversity by integrating features from multiple different levels. Zhang et al.~\citep{zhang2016single} adopt a multi-column CNN architecture, where each column CNN utilizes filters with different receptive fields to adapt to scale variations.
Liu et al.~\citep{liu2019context} propose a novel deep learning architecture that utilizes spatial pyramid pooling to extract context information at different scales. It then uses these scale-aware features to regress the final density map. Ma et al.~\citep{ma2022fusioncount} introduce a multi-scale mechanism that utilizes information provided by different layers in the encoder to generate feature maps with varying receptive field sizes. Han et al.~\citep{han2023steerer} employ a feature selection and inheritance learning approach, along with mask selection and inheritance loss, to perform multi-scale fusion. Wang et al.~\citep{wang2022stnet} use tree-based enhancers to address variations in crowd scale and leverages multi-level auxiliaries to filter out background pixels while adapting to changes in crowd density. These methods achieve good results by effectively integrating multi-scale features. 

Attention mechanisms are also an efficient method. Jiang et al.~\citep{jiang2020attention} adopt a dual-branch structure, where one branch generates adjusted intermediate density maps and scale factors. It then corrects the unevenness of crowd density in the image by combining it with attention masks generated by the other branch, ultimately producing refined attention-based density maps. Lin et al.~\citep{lin2022boosting} employ a learnable global attention module and a local partition attention module, achieving decent performance in counting accuracy. Wang et al.~\citep{wang2023context} adopted a multi-layer context fusion approach and introduced a directive attention mechanism to improve accuracy.


\subsection{Loss Function for Design Crowd Counting}
\label{sec:Loss function design for crowd counting}

Some works have designed loss functions tailored to the characteristics of crowd counting tasks. 
Wang et al.~\citep{wang2020distribution} treated crowd counting as a distribution matching problem, employing Optimal Transport (OT) to find the best match between the generated density map and the point annotations. Yan et al.~\citep{yan2023progressive} proposed a multi-resolution loss function, progressive multi-resolution loss (PML), which can learn to better fit the posterior distribution using multi-scale information, thereby improving counting performance.

\subsection{Other Aspects}
\label{sec:Algorithmic Strategies design for crowd counting}

In addition to model and loss function design, there are also some studies improving crowd counting methods from other different aspects.
Song et al.~\citep{song2021rethinking} developed a novel approach named P2PNet, which directly predicts head coordinates and their confidences in images and matches points to their ground-truth targets using the Hungarian algorithm. Lin et al.~\citep{liu2023point} decompose the pedestrian counting process into point-querying processes and use a custom quad-tree structure to adapt to crowds of different densities. Additionally, it employs a recursive rectangular window attention mechanism to improve computational efficiency.

\section{Method}
\label{method}

\begin{figure*}[ht!]
    \centering
    \makebox[\textwidth][c]{\includegraphics[width=1.07\textwidth]{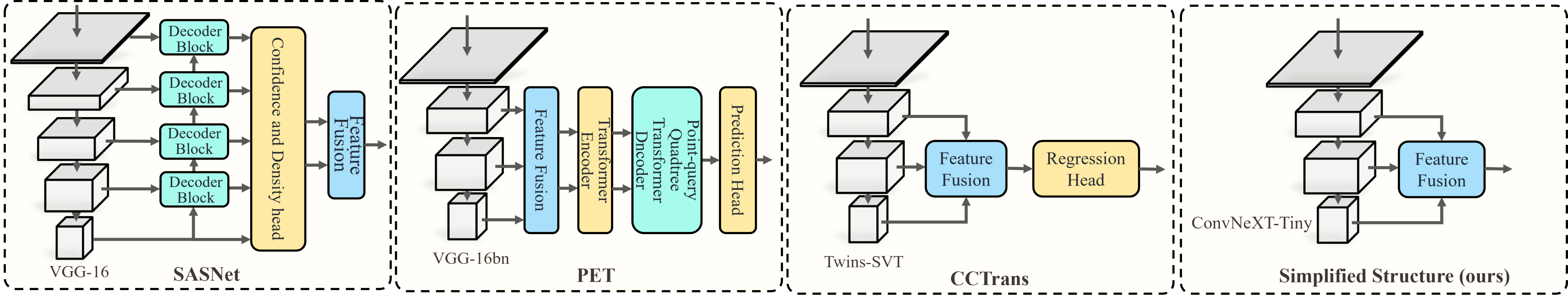}}
    \caption{This figure presents the architectural designs of four different crowd counting models, including SASNet, PET, CCTrans, and our simplified structure. }
    \label{fig:simple architecture}
\end{figure*}

In recent years, a series of high-performance crowd counting models based on deep learning have been proposed. The trade-off for achieving high performance in these models is an increase in the complexity of the model structure. Figure~\ref{fig:simple architecture} illustrates a comparison between the  network structures of three mainstream high-performance crowd counting models, namely CCTrans~\citep{tian2021cctrans}, PET~\citep{liu2023point}, and SASNet~\citep{song2021choose}, and our proposed model structure. More specifically, SASNet incorporates multiple decoder blocks, along with a sophisticated fusion header design and interleaved fusion mechanisms, resulting in a highly intricate architecture.
PET employs the Quadtree structure to address the challenge of crowd counting and incorporates a complex Transformer-based encoder-decoder, substantially amplifying the model's intricacy. Although the structure of CCTrans may seem simple, it actually introduces additional complexity to the overall model. This is because it is built upon the massive parameter Twins-SVT network, even though it only adds a regression header after multi-branch fusion. 

\begin{figure*}[ht]
    \centering
    \includegraphics[width=0.8\textwidth]{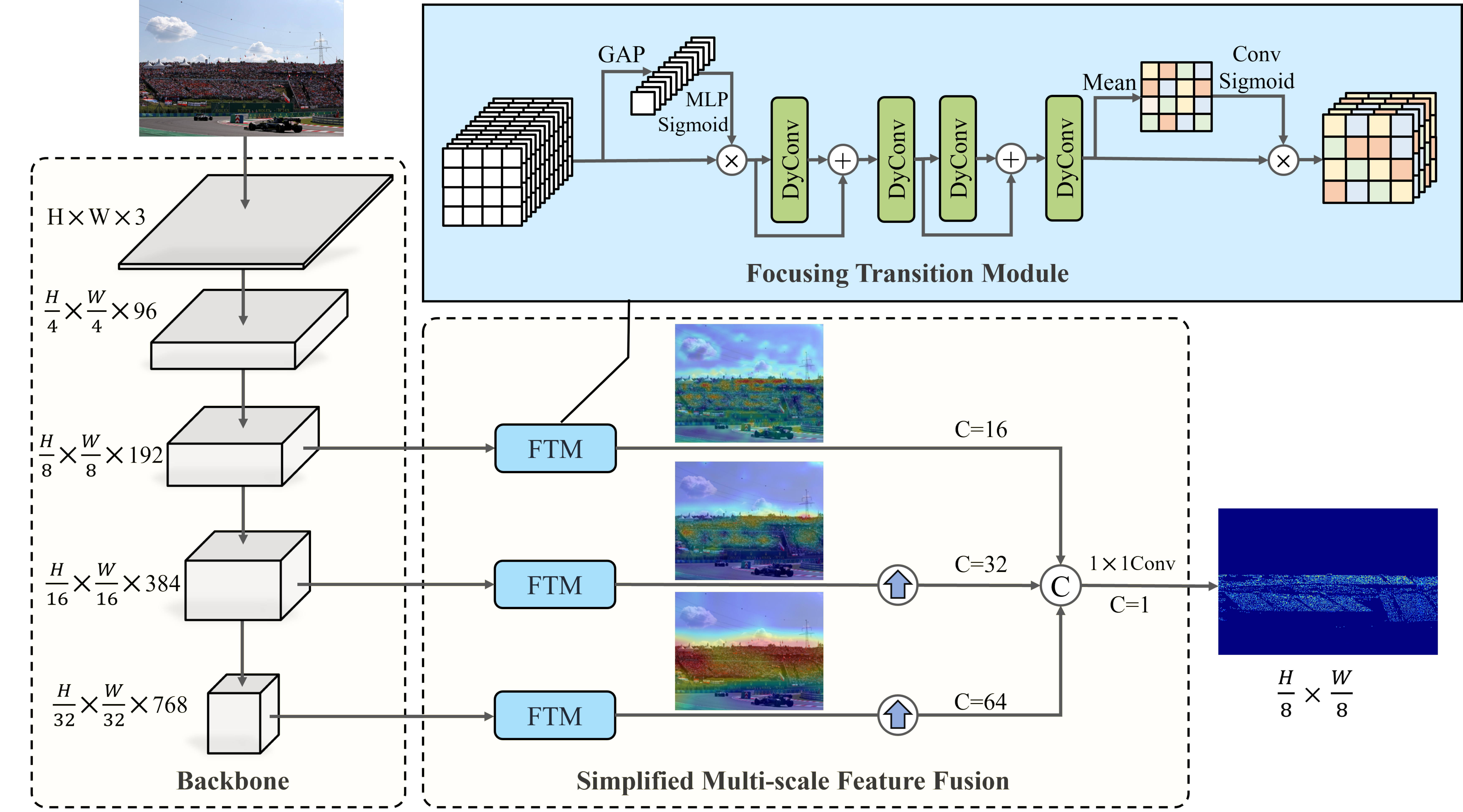}
    \caption{The network architecture of FFNet. $H$, $W$, $C$ denote the height, width, and number of channels of the features, respectively. And the circles with ``×", ``+", and ``C" represent element-wise multiplication, element-wise addition, and channel concatenation operations, respectively.
    }
    \label{fig:FFNet}
\end{figure*}

The fourth structure in Figure~\ref{fig:simple architecture} illustrates the simplified solution proposed by us. We utilize the compact yet powerful ConvNeXT-Tiny as the base backbone and integrate only a multi-scale feature fusion structure with a focus transition module. This model is referred to as Fuss-Free Network (FFNet).
As depicted in Figure \ref{fig:FFNet}, FFNet comprises solely of a backbone and a simplified multi-scale feature fusion module. The backbone is responsible for extracting features at three different scales. The simplified multi-scale feature fusion module includes three branches, each equipped with a focus transition module. The features at three different scales are individually processed through their corresponding focus transition modules, resulting in the generation of more semantically rich features specifically related to crowded scenes. Finally, the features from the three branches are concatenated together, and then processed through a 1×1 convolution layer to merge the features into a single channel, resulting in the final density map. The source code for this project is available on GitHub at \href{https://github.com/erdongsanshi/FFNet}{https://github.com/erdongsanshi/FFNet}.

\subsection{Feature Extraction}
\label{sec:Feature Extraction Backbone}

Given our research objective of developing a crowd counting model with a balanced approach to simplifying the model structure, we opt for ConvNeXT-Tiny~\citep{liu2022convnet} as the backbone due to its notable attributes of simplicity and efficiency. ConvNeXts series build upon the foundation of a standard ResNet~\citep{he2016deep} while drawing inspiration from newer models proposed after 2020, such as vision Transformer and ResNeXt~\citep{xie2017aggregated}, with their innovative design and training techniques, for instance, one can apply techniques such as adjusting the stage compute ratio, utilizing Inverted Bottleneck and Large Kernel Sizes, and substituting ReLU with GELU, among others. By incorporating these operations, ConvNeXt-Tiny is able to preserve the strong spatial processing capabilities of CNNs for local features while also greatly improving the comprehension and utilization of global information.

ConvNeXt-Tiny is a smaller version of the original ConvNeXt model which provides a more efficient computational path and a simplified structure. Its streamlined and efficient design allows it to outperform traditional models in visual tasks like image recognition. This aligns with our goal of creating a concise and efficient network. 
In crowd counting tasks, dealing with objects of various sizes is a major challenge. One common approach is to generate features at different resolutions. To address this, To address this issue, we extract features from diverse stages of ConvNeXt-Tiny, thereby better adapting to the crowd counting task by capturing features at different hierarchical levels and resolutions from each stage.

\subsection{Multi-scale Feature Fusion}
\label{sec:Multi-scale Fusion Strategy}
\begin{figure*}[htb]
    \centering
    \includegraphics[width=1\textwidth]{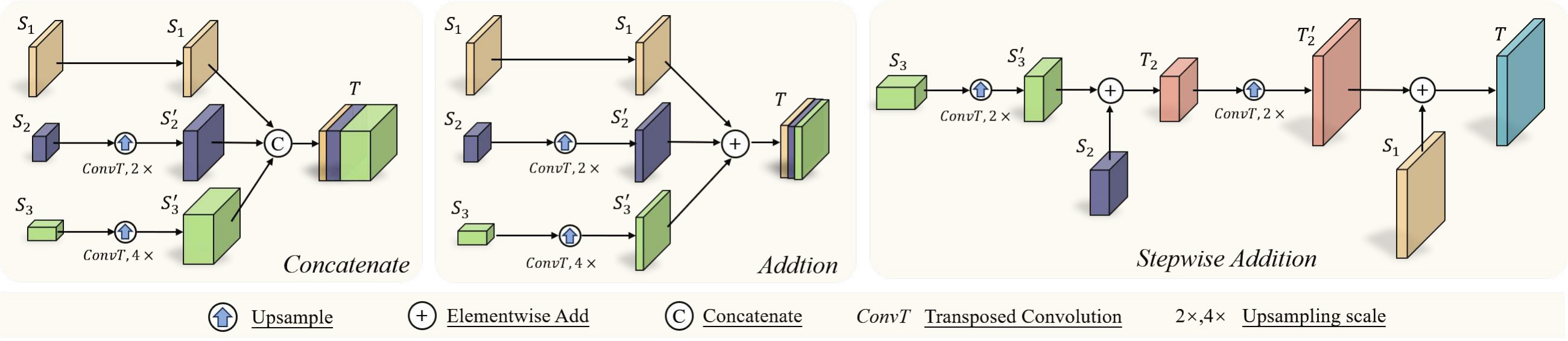}
    \caption{The three simple designs for multi-scale fusion are illustrated below. Given $S_1$, $S_2$, and $S_3$ as inputs, T represents the final output, and $T_2$ represents the output of the $S_2$ stage. }
    \label{fig:Multi-scale_Fusion}
\end{figure*}
The main objective of multi-scale feature fusion is to enhance model performance by effectively integrating features at various scales, which enables the model to capture both fine-grained details and overall contextual information simultaneously. As a result, the model's accuracy, robustness, and generalization capabilities are improved, allowing it to better handle variations in object size, pose, appearance, etc.

In order to adhere to the principle of simplicity and efficiency, we evaluate three commonly used and simple multi-scale feature fusion strategies and select the most efficient one. 
As shown in the Figure~\ref{fig:FFNet}, the ``Concatenate" fusion directly applies transposed convolution to features $S_2$ and $S_3$ for upsampling to match the spatial size of $S_1$, and then fuses the features through channel concatenation. The ``Addition" fusion first reduces the number of channels in features $S_3$ and $S_2$ to match the channel count of $S_1$. Before performing element-wise addition, transposed convolution operations are used to align the spatial size of the features. The ``Stepwise Addition" fusion involves progressively upsampling and combining features of different scales, similar to the design of FPN~\citep{lin2017feature}. 
For the purpose of reconstructing complex crowd distribution features, we specifically select transpose convolution instead of interpolation operations, as it offers notable advantages such as learning capability, flexibility, spatial accuracy, and the ability to handle irregular patterns. This choice allows for more precise and adaptive upsampling, making it particularly impactful in the context of complex crowd distribution scenes.

After conducting experiments, we found that the above three fusion strategies showed satisfactory performance. Among them, the ``Concatenate" fusion strategy yielded the best results. Therefore, we have chosen to adopt this design as the default fusion strategy in the FFNet framework.

\subsection{Focus Transition Module}
\label{sec:Focus Transition Module}

The backbone was originally designed for classification tasks, so the features it outputs contain more general semantic information, which differs from the features required for crowd counting tasks. For example, the most crucial task in crowd counting is to extract features of the crowd. Although the backbone will adapt to crowd counting tasks during the training process, further processing after the three features at different scales will enhance the flexibility of the model, thus better adapting to crowd counting tasks. Furthermore, the direct fusion of the different scales features from backbone may result in information redundancy, potentially reducing learning efficiency. For the aforementioned reasons, we have developed a focus transition module that effectively focuses on dynamic and static features, enabling efficient dimensionality reduction and feature extraction to facilitate the transition. 

According to the aforementioned requirements, we have devised the focus transition module (FTM) to transform the features of the three branches prior to fusion, while also adhering to principles of simplicity and efficiency. 
Figure \ref{fig:FFNet} illustrates the main components of FTM, comprising dynamic convolution based~\citep{li2022omni} feature transformation, a channel attention mechanism, and a spatial attention mechanism. 
To begin with, we utilize a channel attention mechanism to capture the interdependencies among different channels of the feature map extracted from the backbone network. This mechanism assigns weights to each channel according to their significance, enabling the branch to concentrate on pertinent features while suppressing less informative ones. 
Subsequently, the dynamic convolution based feature transformation component is employed to enhance the feature map by effectively capturing both dynamic and static feature variations in the scene. Lastly, we implement a spatial attention mechanism to capture the interdependencies within the spatial of the feature map. This mechanism assigns weights to various spatial locations, allowing the branch to focus on spatial regions while suppressing irrelevant or noisy ones. 

The most crucial component of FTM is the dynamic convolution based feature transformation, which includes dynamic convolution and residual connection. 
Dynamic convolution is employed in our model as an alternative to conventional convolution, effectively enhancing the model's representational power without the need for increasing the depth or width of the network.
Furthermore, we incorporate residual connections throughout the entire process to enhance gradient flow and facilitate the learning of intricate relationships. It is worth noting that while the FTM enhances key features, it also implicitly achieves feature dimensionality reduction. This dimensionality reduction, while preserving the integrity of key information, helpS reduce the computational complexity of the model. In detail, after each dynamic convolution, there is a ReLU activation layer and a BatchNorm layer to ensure the non-linear expression capability of features and the stability of normalization. The entire process of the FTM can be represented by the following formula:  
\begin{align}
C(\mathbf{F})&=\sigma(\operatorname{MLP}(\operatorname{AvgPool}(\mathbf{F}))) \odot \mathbf{F}, \\
P(\mathbf{C})&=Y[Y(Y(Y(\mathbf{C})+\mathbf{C})+(Y(\mathbf{C})+\mathbf{C}))], \\
S(\mathbf{P})&=\sigma(\operatorname{Conv}(\mathbf{\bar{P}})) \odot \mathbf{P},
\label{dyn_equ2}
\end{align}
where $\mathbf{F}$ is the input, $C(\cdot)$ is the channel feature weight optimization, $Y(\cdot)$ is the dynamic convolution, $P(\cdot)$ represents the output after stacked dynamic convolutions, $S(\cdot)$ is the spatial feature weight optimization, $\mathbf{\bar{P}}$ is the mean value of $\mathbf{P}$, and MLP is the multi-layer perceptron.


The fundamental concept behind dynamic convolution is to dynamically combine multiple parallel convolution kernels based on their attentions, which are determined by the input data. This approach generates convolution kernel weights during the runtime of the model, allowing for adaptability to different input data. By capturing features across various dimensions such as input channels, output channels, kernel size, and the number of kernels, dynamic convolution offers flexibility in feature extraction. Dynamic convolution utilizes a combination of average pooling and convolution to determine attention weights, which are then dynamically adjusted. This allows for adaptive parameter adjustment of the convolutional kernel to better align with the input data's features. We define $\mathbf{X}$ as the input of dynamic convolution and the process of dynamic convolution attention extraction and dynamic weighting as follows:
\begin{align}
A(\mathbf{X}) &= \sigma(\operatorname{Conv}(FC(\operatorname{AvgPool}(\textbf{X})))), \\
Y(\mathbf{X}) &= \mathbf{X} \odot \sum_{k=1}^{n}(\mathbf{A}_{n k} \odot \mathbf{A}_{s k} \odot \mathbf{A}_{i k} \odot \mathbf{A}_{o k} \odot \mathbf{W}_{k}),
\label{dyn_equ}
\end{align}
where $A(\cdot)$ represents the attention calculation, AvgPool denotes average pooling, FC stands for fully connected layer, Conv indicates convolution operation, n represents the number of convolutional kernels, and $\sigma$ is the sigmoid activation function; $Y(\cdot)$ represents the dynamic convolution weighting process, $\mathbf{A}_{nk}$, $\mathbf{A}_{sk}$, $\mathbf{A}_{ik}$, $\mathbf{A}_{ok}$ are the attention weights on four dimensions: the number of convolution kernels, kernel size, input channels, and output channels, respectively, $\mathbf{W}_k$ represents the weights of different convolution kernels, and C denotes the input optimized by weights.

\subsection{Loss Function}
\label{sec:Loss Function Design}

We employed the Loss function utilized in DM-Count~\citep{wang2020distribution} which consists of three components: count loss, optimal transport (OT) loss~\citep{ma2019bayesian}, and variation loss.

Count loss is a critical loss in crowd counting that quantifies the gap between the number of people predicted by the model and the actual number of people labeled, thus contributing to better convergence of the model. It is defined as follows:
\begin{align}
\mathcal{L}_{c}(\mathbf{z},\hat{\mathbf{z}})&=\left|\Vert{{\mathbf{z}}}\Vert_1-\Vert{\hat{\mathbf{z}}}\Vert_1\right|,
\label{count loss}
\end{align}
where $\mathbf{z}$ represents the vectorized binary map for dot-annotation and $\hat{\mathbf{z}}$ is the vectorized predicted
density map returned by network, $\Vert \cdot \Vert_1$ denotes the L1 norm.

OTLoss is used to measure the difference between two probability distributions, making it particularly suitable for handling misaligned data distribution problems. It encourages the model to learn probability densities that are closer to the true distribution. It utilizes the Sinkhorn algorithm~\citep{peyre2019computational} for iterative optimization, finding the optimal transportation plan from the source distribution to the target distribution, and obtaining the dual variables during this process. Finally, based on these dual variables and the predicted density distribution, OTLoss is computed. Its definition is as follows:

\begin{align}
\mathcal{L}_{ot}(\hat{\mathbf{z}})&=\langle\frac{\mathbf{\beta}^{\ast}}{\Vert{\hat{\mathbf{z}}}\Vert_1}-\frac{\langle \mathbf{\beta}^{\ast}, \hat{\mathbf{z}} \rangle}{\Vert{\hat{\mathbf{z}}}\Vert^2_1}, \hat{\mathbf{z}}\rangle,
\label{Distance matrix D}
\end{align}
where $\beta^{\ast}$ is  the solutions of OT cost~\citep{wang2020distribution}, which are computed using the Sinkhorn algorithm;  $\langle\cdot,\cdot\rangle$ represents the dot product of two vectors.

Furthermore, variation loss is utilized to measure the intensity or grayscale variation differences between adjacent pixels in the predicted and ground truth density maps. Specifically, the L1 norm is employed. We utilize this loss function to improve the overall image smoothness and effectively minimize the negative impact of potential noise on density estimation. To balance the contribution of losses from areas with different crowd densities, we use the actual crowd count as a weighting factor to adjust the variance loss. Its definition is as follows:
\begin{align}
\mathcal{L}_{v}(\mathbf{z},\hat{\mathbf{z}})&=\Vert{{\mathbf{z}}}\Vert_1 \cdot \Vert|\mathbf{z}-\hat{\mathbf{z}}\Vert_1.
\label{variation loss}
\end{align}

The total loss is the weighted sum of three components, which is defined as:
\begin{align}
\mathcal{L}_{all}&=\mathcal{L}_c+\lambda_1\mathcal{L}_{ot}+\lambda_2\mathcal{L}_{v},
\label{loss}
\end{align}
where $\lambda_1$ and $\lambda_2$ are the weighting values, and  were assigned as 0.1 and 0.01, respectively, in the experiment.

\section{Experiments}
\label{Experiment}


\subsection{Implementation Details}
\label{sec:Implementation Details}

\subsubsection{Dataset}
\label{sec:Dataset}

We evaluate our methods in four benchmarks, including UCF\textunderscore CC\textunderscore50~\citep{idrees2013multi}, ShanghaiTech Part A (SHA) and ShanghaiTech Part B (SHB)~\citep{zhang2016single}, and NWPU-Crowd~\citep{wang2020nwpu}. The datasets, described in detail in Table\ref{tab:datasets}, vary in terms of scene complexity, crowd density range, and image resolution.

\begin{table*}[htbp]
    \centering
    \begin{adjustbox}{width=1\textwidth, center}
    \begin{tabular}{lllllll}
    \toprule
    Datasets & Numbers of Image & Average Resolution & Total marked persons &  Max marked persons & Min marked persons & Average  marked persons  \\ \hline
    UCF\textunderscore CC\textunderscore50 & 50 & 2101 × 2888 & 63974 & 4633 & 94 & 1279 \\
    ShanghaiTech A(SHA) & 482 & 589 × 868 & 241677 & 3139 & 33 & 501 \\
    ShanghaiTech B(SHB) & 716 & 768 × 1024 & 88488 & 578 & 9 & 123  \\
      NWPU & 5109 & 2191 × 3209 & 2133375 & 20033 & 0 & 418 \\ 
    \bottomrule
    \end{tabular}
    \end{adjustbox}
    \caption{
    Crowd counting datasets used for evaluation.
    }
    \label{tab:datasets}
\end{table*}

\subsubsection{Training Settings and Hyperparameters}
\label{sec:Training Settings and Hyperparameters}

For a fair comparison, the backbone of FFNet is the official ConvNeXt-Tiny model, which is pretrained on the ImageNet-1k dataset. In our experiments, data augmentation was limited to random cropping and random horizontal flipping. For SHB, we set the cropping size to 512; for NWPU, the cropping size is set to 384; and for UCF\textunderscore CC\textunderscore50 and SHA datasets, we adjust the cropping size to 256. The model training utilized the AdamW optimizer with a batch size of 8, and the initial learning rate was set to 1e-5. Additionally, to mitigate overfitting, we introduced an L2 regularization term with a coefficient of 0.005. All experiments were conducted using PyTorch on a single 16-GB RTX 4080.

\subsubsection{Evaluation Metrics}
\label{sec:Evaluation Metrics}

In crowd counting research, Mean Absolute Error (MAE) and Mean Squared Error (MSE) are commonly used as evaluation metrics for crowd counting.
The MAE and MSE are defined as follows:
\begin{align}
M A E&=\frac{1}{N} \sum_{i=1}^N\left|\mathbf{c}_i-\hat{\mathbf{c}}_i\right|, \\
M S E&=\sqrt{\frac{1}{N} \sum_{i=1}^N\left(\mathbf{c}_i-\hat{\mathbf{c}}_i\right)^2},
\label{MAE,MSE}
\end{align}
where $N$ is the number of test images, and $\mathbf{c}_i$ and $\hat{\mathbf{c}}_i$ are the  true and predicted number of people in the $i$-th image, respectively.

\begin{table*}[htbp]
    \centering
    \scriptsize
    \begin{adjustbox}{width=1.05\textwidth, center}
    \begin{tabular}{lllllllllllll}
    \toprule
    \multirow{2}{*}{Method} & \multirow{2}{*}{Venue} & \multirow{2}{*}{Backbone} & \multirow{2}{*}{Parameters} & \multirow{2}{*}{FLOPS} & \multicolumn{2}{c}{UCF\textunderscore CC\textunderscore50} & \multicolumn{2}{c}{SHA} & \multicolumn{2}{c}{SHB} & \multicolumn{2}{c}{NWPU(V)} \\ \cline{6-13}    
        &  &  &  &  & MAE & MSE & MAE & MSE & MAE & MSE & MAE & MSE \\ \hline
            CAN~\citep{liu2019context} & CVPR19 & VGG-16 &-&-& 212.2 & 243.7 & 62.3 & 100.0 & 7.8 & 12.2 & 93.5 & 489.9 \\
      ASNet~\citep{jiang2020attention} & CVPR20 & VGG-16 &-&-& 174.8 & 251.6 & 57.8 & 90.1 & - & - & - & - \\
        DensityCNN~\citep{jiang2020density} & TMM20 & VGG-16 &-&-& 244.6 & 341.8 & 63.1 & 106.3 & 9.1 & 16.3 & - & - \\
 DM-Count~\citep{wang2020distribution} & NeurIPS20 & VGG-19 &-&-& 211.0 & 291.5 & 59.7 & 95.7 & 7.4 & 11.8 & 70.5 & 357.6 \\
     P2PNet~\citep{song2021rethinking} & ICCV21 & VGG-16 & \underline{19.22M} & 104.71G & 172.7 & 256.2 & 52.7 & 85.1 & 6.3 & 9.9 & - & - \\
         DKPNet~\citep{Chen_2021_ICCV} & ICCV21 & HRNet-W40 & \textbf{12.60M} & 46.31G & - & - & 55.6 & 91.0 & 6.6 & 10.9 & 61.8 & 438.7 \\
         SASNet~\citep{song2021choose} & AAAI21 & VGG-16 & 38.90M & 232.98G & \underline{161.4} & 234.46 & 53.5 & 88.3 & 6.3 & 9.9 & - & - \\
         PFDNet~\citep{yan2021crowd} & TMM21 & VGG-16 &-&-& 205.8 & 289.3 & 53.8 & 89.2 & 6.5 & 10.7 & - & - \\
           MAN~\citep{lin2022boosting} & CVPR22 & VGG-19 & 30.96M & 105.85G & - & - & 56.8 & 90.3 & - & - & - & - \\
             CHFL~\citep{shu2022crowd} & CVPR22 & VGG-19 & 21.51M & 108.27G & - & - & 57.5 & 94.3 & 6.9 & 11.0 & 76.8 & 343.0 \\
             CLTR~\citep{liang2022end} & ECCV22 & ResNet-50 & 40.95M & \underline{28.73G} & - & - & 56.9 & 95.2 & 6.5 & 10.6 & 61.9 & 246.3 \\
           FIDT~\citep{liang2022focal} & TMM22  & HRNet-W48 & 66.58M & 142.40G & 171.4 & \textbf{233.1} & 57.0 & 103.4 & 6.9 & 11.8 & 51.4 & \underline{107.6}\\
        PML~\citep{yan2023progressive} & TCSVT23 & ConvNeXt-S &-&-& - & - & 53.4 & 93.1 & 6.2 & 9.7 & - & - \\
              PET~\citep{liu2023point} & ICCV23 & VGG-16 & 51.65M & 859.68G & - & - & \underline{49.3} & \textbf{78.9} & 6.2 & 9.7 & 58.5 & 238.0 \\
        STEERER~\citep{han2023steerer} & ICCV23 & HRNet-W48 & 64.42M & 94.40G & - & - & 54.5 & 86.9 & \textbf{5.8} & \textbf{8.5} & 54.3 & 238.3 \\ \hline
          BCCT~\citep{sun2021boosting} & arXiv21 & T2T-ViT-14 &-&-& - & - & 53.1 & 82.2 & 7.3 & 11.3 & 53.0 & 170.3 \\
       CCTrans~\citep{tian2021cctrans} & arXiv21 & Twins-SVT & 103.58M & 99.44G & 168.7 & 234.5 & 52.3 & 84.9 & 6.2 & 9.9 & \textbf{38.6} & \textbf{87.8} \\
      CUT~\citep{qian2022segmentation} & BMVC22 & Twins-PCPVT & 68.09M & 90.45G & - & - & 51.9 & \underline{79.1} & - & - & - & - \\ \hline
     FFNet(ours) & - & ConvNeXt-T & 29.02M & \textbf{23.67G} & \textbf{161.1} & \underline{233.3} & \textbf{48.3} & 80.5 & \underline{6.1} & \underline{9.0} & \underline{41.2} & 113.2 \\
\bottomrule
\end{tabular}
\end{adjustbox}
\caption{
Compared with the advanced methods on four datasets. 
(The best results are shown in \textbf{bold}, and the second-best results are \underline{underlined}).}
\label{tab:table1}
\end{table*}

\subsection{Comparison with state-of-the-art methods}
\label{Comparison with state-of-the-art methods}

\subsubsection{Counting performance}
\label{sec:Counting performance}

CNN-based crowd counting methods dominate the field of crowd counting, with most approaches building upon the foundation of CNNs for model design. As shown in Table \ref{tab:table1}, our method outperforms them, exhibiting significant advantages in several benchmark tests and achieving SOTA or near-state-of-the-art performance levels. Even when compared to the popular transformer-based methods in recent years, our model demonstrates superiority, achieving better results in most cases.

Specifically, we achieved outstanding results on the UCF\textunderscore CC\textunderscore50 dataset, with an MAE of 161.4, surpassing existing methods and reaching a new SOTA level. On the SHA subset, we demonstrated significant superiority with an MAE value of 48.3, successfully surpassing the latest SOTA in this field. In the evaluation of the SHB subset, despite fierce competition, our method still achieved a highly competitive second-place performance with an MAE of 6.1. Finally, in the more challenging NWPU-Crowd dataset, our method also achieved results close to the top tier. Without excessive complexity in design, our model achieves excellent results on these datasets.

\subsubsection{Parameters and Computational Resources}
\label{sec:Lightweight}

Adhering to the original intention of our work, we aimed to simplify the model while achieving high crowd counting performance. Therefore, we conducted lightweight testing on FFNet and SOTA methods. The testing primarily focused on open-source models released in the past three years. We did not test some methods that are partially open-source or not open-source. For a fair comparison, we all test with an all-one tensor of size (1, 3, 512, 512) as input. And we measured the lightweight characteristics of the model based on two metrics: model parameters and floating-point operations (FLOPs). Among them, the model parameters is one of the key indicators for measuring the complexity of neural network models. It refers to the total number of weights and bias parameters in the model. Models with fewer parameters occupy less storage resources, have faster inference speeds, are easier to deploy, and have lower risks of overfitting. FLOPS is another important metric for measuring the complexity and computational requirements of neural network models. In the experiment, it refers to the number of floating-point operations required for one forward pass. Models with lower FLOPS consume fewer computational resources, have faster inference speeds, are more suitable for low-power environments, and are easier to deploy on mobile and embedded devices.

The experimental testing results are presented in Table \ref{tab:table1}. The data shows that over the past three years, as counting performance has continued to improve, there has been a significant increase in model parameters and FLOPs, resulting in greater model complexity. However, FFNet's parameters and FLOPs are significantly lower than those of other advanced crowd counting methods. While achieving superior counting performance, achieve optimal FLOPs and highly competitive model parameters. This effectively proves the efficacy of our proposed model: concise, efficient, low in complexity, and yet powerful in performance. FFNet effectively balances model simplification and performance enhancement, making it suitable not only for deployment in resource-constrained environments but also for performing exceptionally well in scenarios with high real-time requirements. 

\subsection{The Effectiveness of Our Simplified Structure}
\label{sec:The Effectiveness of Simple Network}

To verify the effectiveness of our concise structure, we conducted replacement experiments on the backbone of the structure. 
We replaced the backbone of our model, CCTrans, and PET, with different pre-trained networks. Subsequently, we performed parameter fine-tuning on the modified models mentioned above using the SHA dataset, and conducted a count of the number of parameters and Flops for each model after replacing the backbone, using a 512×512 input. The results are shown in Table \ref{tab:replace_backbone}. 
When using ConvNeXt-Tiny and Swin-Transformer-Small, our structure achieved significantly lower MAE and MSE with a minimal number of parameters and floating-point operations compared to the other CCTrans and PET.
In addition, when using MobileNet-v3 as the backbone, our structure achieved MAE and MSE that are second only to CCTrans, with the lowest number of parameters (3.05M) and floating-point operations (1.51G).
The experimental results show that our structure exhibits better robustness and  effectiveness with lower parameter and computational complexity compared to other the structures of CCTrans and PET. 
\begin{table*}[htbp]
    \centering
    \begin{adjustbox}{width=1\textwidth, center}
    \begin{tabular}{lllllllllllll}
    \toprule
    \multirow{2}{*}{Method} & \multicolumn{4}{c}{MobileNet-v3} & \multicolumn{4}{c}{ConvNeXt-Tiny} & \multicolumn{4}{c}{Swin-Transformer-Small}       \\ \cline{2-13} 
    & \multicolumn{1}{l}{Parameters} & FLOPS & MAE & MSE & \multicolumn{1}{l}{Parameters} & FLOPS & MAE & MSE & \multicolumn{1}{l}{Parameters} & FLOPS & MAE & MSE \\ \hline
    
     CCTrans~\citep{tian2021cctrans} & 4.74M & 9.08G & \textbf{62.9} & \textbf{102.1} & 32.12M & 41.01G & 53.9 & 86.5 & 37.35M & 47.78G & 56.6 & 92.4 \\
            PET~\citep{liu2023point} & 15.19M & 133.81G & 66.6 & 110.6 & 90.96M & 343.54G & 54.1 & 85.1 & 106.68M & 404.46G & 58.1 & 90.0 \\ \hline
        FFNet & \textbf{3.05M} & \textbf{1.51G} & 64.3 & 109.3 & \textbf{29.02M} & \textbf{23.67G} & \textbf{48.3} & \textbf{80.5} & \textbf{34.26M} & \textbf{30.44G} & \textbf{54.7} & \textbf{88.1} \\
    \bottomrule
    \end{tabular}
    \end{adjustbox}
    \setlength{\abovecaptionskip}{0.1cm}
    \caption{Backbone replacement experiment. 
    (Comparision of the effectiveness of different structs.)
    }
    \label{tab:replace_backbone}
\end{table*}
We also compare our model with recent lightweight crowd counting models using MobileNet-v3 as the backbone. As show in Table~\ref{tab:mobile_compare}, by simply using a lightweight backbone, our model easily achieves lightweight performance.
\begin{table}[htbp]
    \centering

    \begin{tabular}{lllll}
    \toprule
    \multirow{2}{*}{Method} & \multirow{2}{*}{Parameters} & \multirow{2}{*}{FLOPS(input size)} & \multicolumn{2}{c}{SHA}       \\ \cline{4-5} 
    & & & \multicolumn{1}{l}{MAE} & MSE \\ \hline
    ECCNAS-Lat~\citep{wang2022eccnas} & 3.79M & 7.38G(512×512) & 66.3 & 116.5 \\
      ECCNAS-Acc~\citep{wang2022eccnas} & 3.88M & 16.46G(512×512) & 62.0 & 110.9\\
      LMSFFNet~\citep{yi2023lightweight} & 4.58M & 14.9G(1024×768) & 85.8 & 139.96\\ \hline
        FFNet & \multirow{2}{*}{3.05M} & 1.51G(512×512) & \multirow{2}{*}{64.3} & \multirow{2}{*}{109.3} \\
       (MobileNet-v3) &       & 4.53G(1024×768) &     & \\
    \bottomrule
    \end{tabular}
    \setlength{\abovecaptionskip}{0.1cm}
    \caption{
    Comparison with advanced lightweight models. (We utilized 256×256 cropped images to calculate the MAE and MSE metrics for the models.)
    }
    \label{tab:mobile_compare}
\end{table}

\subsection{Multi-scale feature fusion}
\label{sec:Multi-scale Feature fusion}

In this section, we conducted comparative experiments on three simple multi-scale fusion strategies in Figure \ref{fig:Multi-scale_Fusion}. The experiment only adjusted the fusion method of the FFNet and validated it on the SHA and SHB datasets. The experimental results are shown in Table \ref{tab:Fusion in FFNet} below. In SHA, both the ``Addition” and ``Stepwise addition" operations demonstrate good performance, but the ``Concatenate" operation outperforms them. In SHB, the ``Concatenate" operation shows better performance and robustness, which proves the effectiveness of the ``Concatenate" operation in crowd counting tasks.
\begin{table}[htbp]
    \centering
    \begin{tabular}{lllll}
    \toprule
    \multirow{2}{*}{Fusion Strategy} & \multicolumn{2}{c}{SHA} & \multicolumn{2}{c}{SHB}       \\ \cline{2-5} 
                                     & \multicolumn{1}{l}{MAE} & MSE & MAE & MSE   \\ \hline
                          Addition  & \multicolumn{1}{l}{52.6} & 87.5 & 6.4 & 10.8     \\
            Stepwise addition  & \multicolumn{1}{l}{52.3} & 91.9 & 7.0 & 11.7   \\ 
                  Concatenate  & \multicolumn{1}{l}{\textbf{48.3}} & \textbf{80.5} & \textbf{6.1} & \textbf{9.0}\\ 
\bottomrule
\end{tabular}
\setlength{\abovecaptionskip}{0.1cm}
\caption{Comparison of three simple multi-scale fusion designs on SHA and B datasets.}
\label{tab:Fusion in FFNet}
\end{table}

\subsection{Ablation Study}
\label{sec:Ablation Study}

As mentioned earlier, our proposed FFNet consists of the following components: a feature extraction backbone, focus transition modules, and a multi-scale fusion structure. We conducted extensive ablation experiments on the SHA dataset to validate the effectiveness of each component in FFNet.

\subsubsection{Focus Transition Module}
\label{sec:FTM}
\begin{table}[htbp]
    \centering
    \scriptsize
    \begin{tabular}{lll}
    \toprule
    \multirow{2}{*}{Method} & \multicolumn{2}{c}{SHA}        \\ \cline{2-3} 
                                 & \multicolumn{1}{l}{MAE} & MSE \\ \hline
        F1$\rightarrow$F1+FTM & 76.3$\rightarrow$\textbf{72.1} & 132.2$\rightarrow$\textbf{127.0} \\
        F2$\rightarrow$F2+FTM & 55.6$\rightarrow$\textbf{54.1} & 95.7$\rightarrow$\textbf{93.1} \\
        F3$\rightarrow$F3+FTM & 54.5$\rightarrow$\textbf{50.9} & 93.1$\rightarrow$\textbf{84.4} \\
        F1+F2+F3(Without FTM)$\rightarrow$F1+F2+F3(With FTM) & 54.2$\rightarrow$\textbf{48.3} & 91.7$\rightarrow$\textbf{80.5} \\
    \bottomrule
    \end{tabular}
    \caption{Effective improvement of crowd counting performance through FTM.}
    \label{tab:FTM}
\end{table}
We conducted two ablation experiments on the FTM module, one focusing on local effects and the other on global effects, to verify its effectiveness. First, at the local level, we tested the counting results of the three branches of the model before simple fusion on dataset A. Then, based on this, we removed the FTM module and tested the counting results of the three branches again, and compared the results. The effectiveness of FTM will demonstrate by observing the difference of counting performance before and after the three branches. At the global level, we tested the performance of FFNet with and without FTM to highlight the overall contribution of FTM to FFNet. The results are shown in Table \ref{tab:FTM}. The results indicate that employing the FTM module, whether in local branches or in the overall model, effectively improves the counting accuracy, demonstrating the practicality of the focus transition module.

\subsubsection{Component Analysis of FFNet}
\label{sec:Composition Analysis of FFNet}
\begin{table}[htbp]
    \centering
    \begin{tabular}{lll}
    \toprule
    \multirow{2}{*}{Method} & \multicolumn{2}{c}{SHA}        \\ \cline{2-3} 
                                 & \multicolumn{1}{l}{MAE} & MSE \\ \hline
        backbone & 54.5 & 93.1 \\
        backbone+FTM & 50.9 & 84.4 \\
        backbone+FTM+Concatenate & \textbf{48.3} & \textbf{80.5} \\
    \bottomrule
    \end{tabular}
    \caption{Component analysis of of FFNet.}
    \label{tab:Composition Analysis of FFNet}
\end{table}
Table \ref{tab:Composition Analysis of FFNet} displays the component analysis of our model. For the SHA dataset, both FTM and the final ``Concatenate" fusion contribute significantly to the final MAE metric.

\subsection{Effective Receptive Field and Heatmap}
\label{sec:Visualization}
\begin{figure}[htbp]
    \centering
    \includegraphics[width=0.45\textwidth]{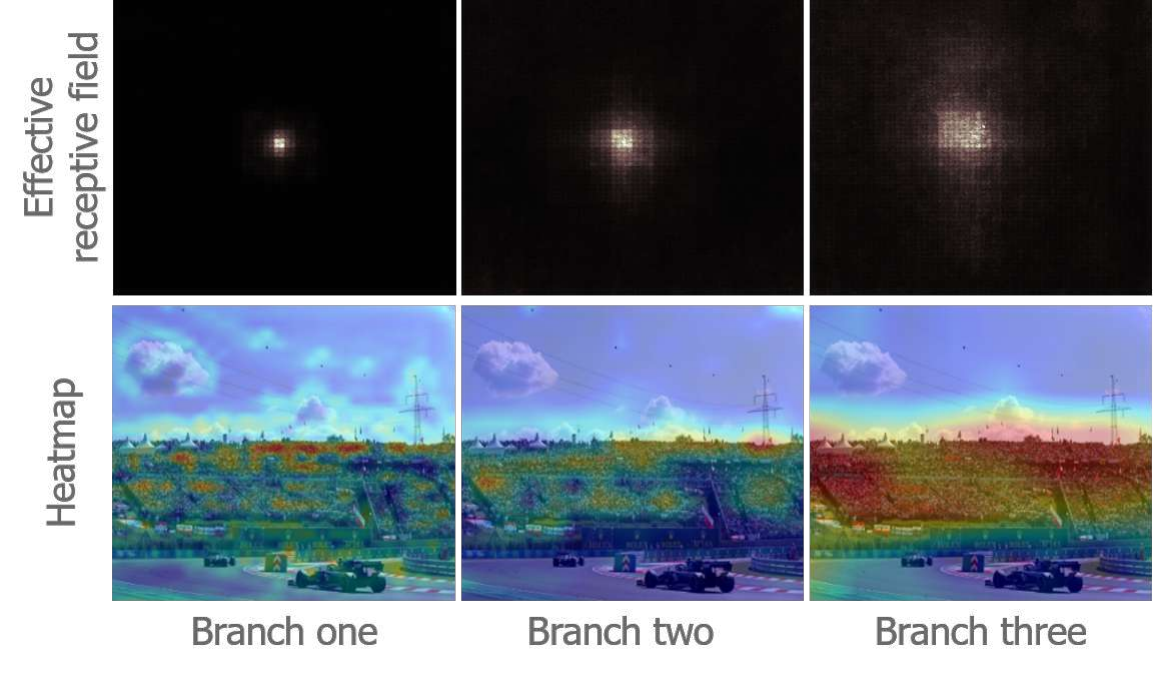}
    \caption{Three branches of the model pass FTM module in the receptive field map and their heat map.}
    \label{fig:visual1}
\end{figure}
To enhance the intuitive understanding of our model's internal mechanism, we visualized the effective receptive field (ERF)~\citep{luo2016understanding} as well as the heat maps of the three branches before fusion. This allows us to uncover the spatial coverage and response intensity distribution of the input data features in each branch before the fusion occurs. 

The theoretical Receptive Field (RF) denotes the area in the input space that directly influences the output of an individual neuron in a neural network. However, the contribution of each pixel within a receptive field to the response of an output unit may not be equal. The effective receptive field is employed to quantify the extent to which each pixel on the input feature map affects a pixel on the output feature map. Typically, the impact within a receptive field follows a Gaussian distribution. The heatmap is a visualization tool that represents numerical intensity through color coding and is used to display the model's level of attention or confidence distribution towards specific targets or features in the input data. 

As shown in Figure \ref{fig:visual1}, the first row is the effective receptive field of three different branches, the second row is the corresponding heat map and each column represents a different branch of the FFNet. From the effective receptive field maps, we can see that the effective receptive field of different branches is different, and the deeper the branch, the larger the effective receptive field, which indicates that our model can capture information at different scales through three branches. The heatmap shows the degree of attention of each branch to the input image through the change in color shade. The first branch is more concerned with the low-level, local, and basic visual information of the image, such as edge, texture, color, etc., and has good maintenance of spatial details. In the second and third branches, attention gradually increases, focusing on abstract and macroscopic features. The color distribution of the heatmaps also becomes more concentrated, demonstrating the model's ability to integrate global information and recognize the overall distribution and density of the crowd.

\begin{figure*}[htbp!]
    \centering
    \includegraphics[width=1\textwidth]{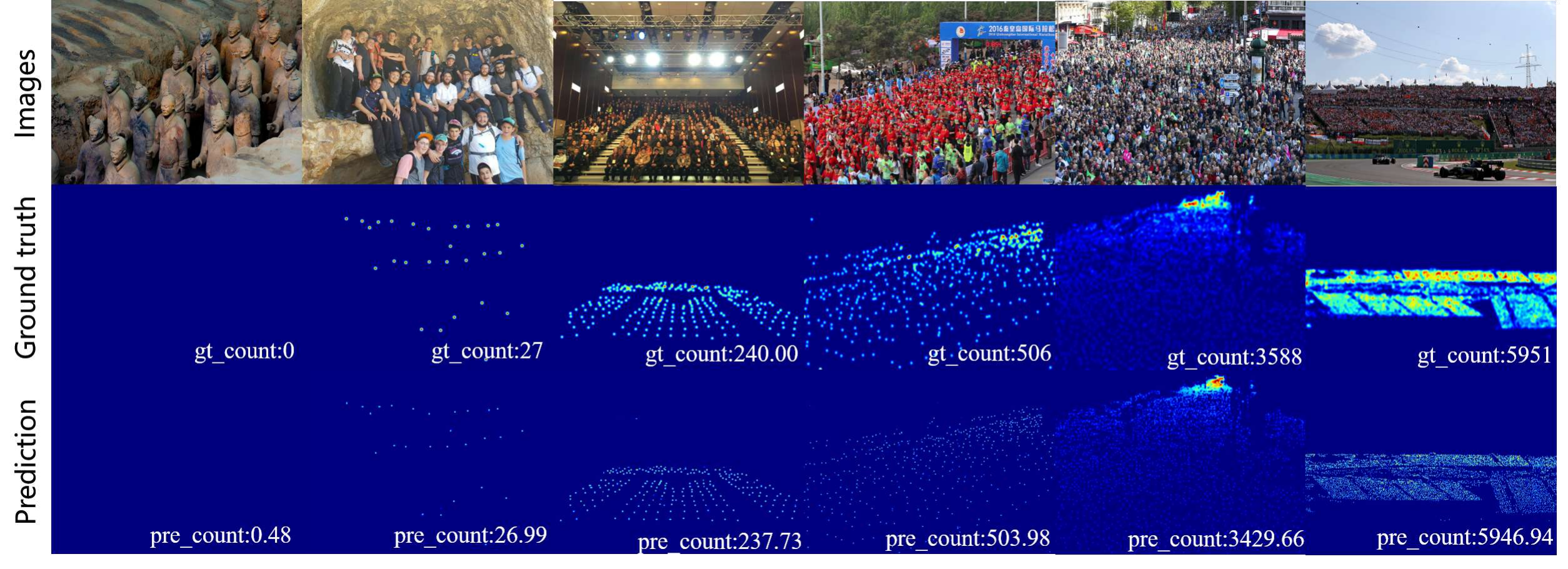}
    \caption{FFNet visualizations on the most challenging NWPR populations. Six images from different scenes, with different scales, densities, and lighting differences, show the counting and fitting capabilities of FFNet.}
    \label{fig:visual2}
\end{figure*}

Finally, we visualize the model outputs using six images in the NWPU dataset, which exhibit varying scales, densities, lighting conditions and from different scenes. The visualization of the model's final outputs is presented in Figure \ref{fig:visual2}. Our model performs excellently on images of various scales, demonstrating good the fitting ability.

\section{Conclusion}
\label{Conclusion}

In this paper, we propose a simplified and efficient crowd counting model called Fuss-Free Network (FFNet), which consists only of a backbone network, a multi-scale fusion structure with focus transition modules. We use the ConvNeXt-Tiny as backbone network and build a simple feature fusion that contains three branches. Each branch is equipped with its focus transition module, which has the capability to enhance the model's flexibility and extract a greater variety of features related to the crowd. The features from three branches are fused through a concatenate operation and a $1\times1$ convolution is applied to generate a single-channel density map. The experiment results show that the counting accuracy of the model is equal to or even better than that of the complex models, despite its compact structure, small number of parameters, and low computational complexity. 
FFNet provides a simple and efficient solution for crowd counting tasks, but there is still plenty of room to explore. For example, cross-domain adaptability, interpretability, and robustness are improved. In our future work, we will explore the field of crowd counting from the perspectives of simplicity, robustness, and high counting accuracy, hoping to build a more general, reliable, and robust model. 


\ifCLASSOPTIONcaptionsoff
  \newpage
\fi



%
\small
\bibliographystyle{IEEEtran}
\end{document}